\documentclass[10pt,twocolumn,letterpaper]{article}

\usepackage{cvpr}              % To produce the CAMERA-READY version
\usepackage[dvipsnames]{xcolor}

\usepackage{amsmath}
\usepackage{amssymb}
\usepackage{booktabs}
\usepackage{dsfont}
\usepackage{epigraph}
\usepackage{graphicx}
\usepackage{lilyglyphs}

\definecolor{cvprblue}{rgb}{0.21,0.49,0.74}
\usepackage[pagebackref,breaklinks,colorlinks,citecolor=cvprblue]{hyperref}

\def\paperID{SynData4CV} % *** Enter the Paper ID here
\def\confName{CVPR}
\def\confYear{2024}

\title{Paved2Paradise: Cost-Effective and Scalable LiDAR Simulation by Factoring the Real World}

\author{Michael A. Alcorn$^{1}$\thanks{Corresponding author: \texttt{michael@bearflagrobotics.com}.} \quad Noah Schwartz$^{1,2}$\\
$^{1}$Bear Flag Robotics\\$^{2}$ John Deere
}

\begin{document}
\maketitle

\begin{abstract}
To achieve strong real world performance, neural networks must be trained on large, diverse datasets; however, obtaining and annotating such datasets is costly and time-consuming, particularly for 3D point clouds.
In this paper, we describe Paved2Paradise\footnote{Code for the Paved2Paradise pipeline can be found at: 
\url{https://github.com/airalcorn2/paved2paradise}
% $<$anonymized$>$
.}, a simple, cost-effective approach for generating fully labeled, diverse, and realistic lidar datasets \textit{from scratch}—all while requiring minimal human annotation.
Our key insight is that, by deliberately collecting \textit{separate} ``background'' and ``object'' datasets (i.e., ``factoring the real world''), we can intelligently combine them to produce a combinatorially large and diverse training set.
The Paved2Paradise pipeline thus consists of four steps: (1) collecting copious background data, (2) recording individuals from the desired object class(es) performing different behaviors in an isolated environment (like a parking lot), (3) bootstrapping labels for the object dataset, and (4) generating samples by placing objects at arbitrary locations in backgrounds.
To demonstrate the utility of Paved2Paradise, we generated synthetic datasets for two tasks: (1) human detection in orchards\footnote{The name ``Paved2Paradise'' refers to the fact that we move human point clouds from parking lot scenes (i.e., paved environments) to orchard backgrounds (i.e., ``paradises'' like the Garden of Eden) for this task.} (a task for which no public data exists) and (2) pedestrian detection in urban environments.
Qualitatively, we find that a model trained exclusively on Paved2Paradise synthetic data is highly effective at detecting humans in orchards, including when individuals are heavily occluded by tree branches.
Quantitatively, a model trained on Paved2Paradise data that sources backgrounds from KITTI performs comparably to a model trained on the actual dataset.
These results suggest the Paved2Paradise synthetic data pipeline can help accelerate point cloud model development in sectors where acquiring lidar datasets has previously been cost-prohibitive.
\end{abstract}

\section{Introduction and Related Work}\label{sec:intro}

\begin{figure*}[htbp]
\centering
\includegraphics[scale=0.25]{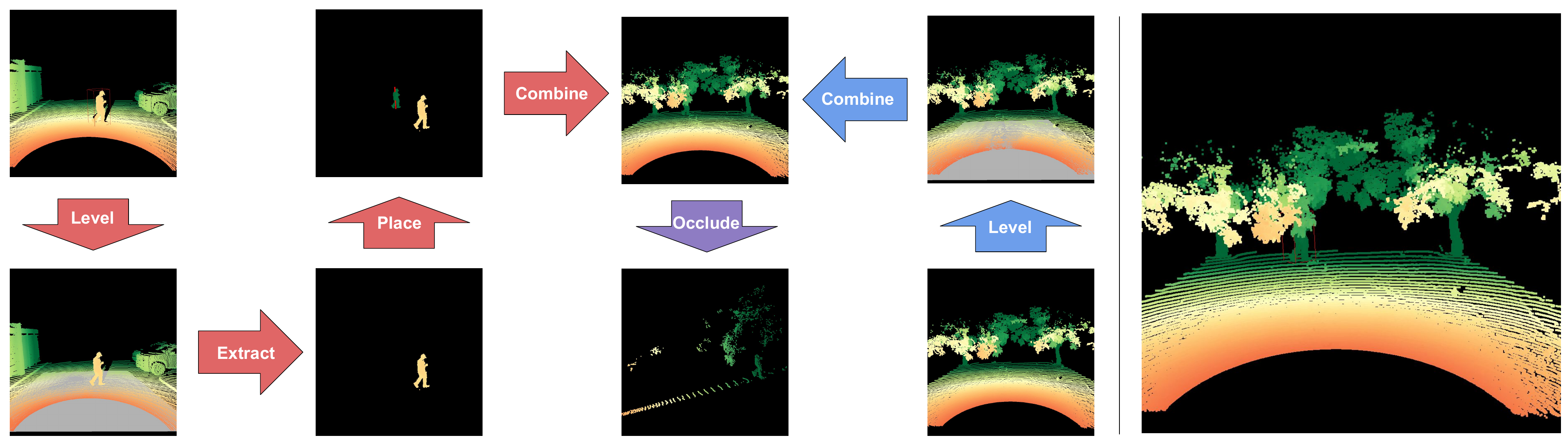}
\setlength{\belowcaptionskip}{-16pt}
\caption{An overview of the Paved2Paradise pipeline for generating synthetic scenes (left) and an example synthetic scene (right).
\textbf{Left}: The pipeline begins by randomly sampling an object scene (top left) and a background scene (bottom right) and ``leveling'' both scenes (Section \ref{sec:level}).
The leveling step ensures the object point cloud extracted from the object scene will be placed on the ground in the background scene.
Next, the object point cloud is extracted from the object scene and placed at a randomly sampled location in the detection region (Section \ref{sec:place}).
During the placement step, the object point cloud is rotated to ensure perspective consistency relative to the sensor's location.
The repositioned object point cloud is then combined with the background point cloud (Section \ref{sec:combine}), and a final occluding procedure removes points from the scene based on their visibility from the sensor's perspective (Section \ref{sec:occlude}).
\textbf{Right}: The resulting synthetic scene is both realistic and automatically annotated.}
\label{fig:overview}
\end{figure*}

To achieve strong real world performance, neural networks must be trained on large, diverse datasets; however, obtaining and annotating such datasets is costly and time-consuming.
For 3D object detection and semantic segmentation on point clouds, the annotation process is particularly slow due to the inherent challenges of working in 3D environments (e.g., confusing perspectives) \cite{behley2021towards}.
As a result, most lidar benchmark datasets are quite small by modern deep learning standards, particularly when compared to image datasets like ImageNet \cite{deng2009imagenet} and COCO \cite{lin2014microsoft}.
The KITTI \cite{geiger2012we, geiger2013vision}, SemanticKITTI \cite{behley2019semantickitti}, and nuScenes \cite{caesar2020nuscenes} datasets contain 7,481, 23,201, and 40,157 annotated frames, respectively.
Further, the frames for each of these datasets were collected in short sequences, i.e., the frames in a sequence are highly correlated, and thus the environmental variability reflected in each dataset is considerably lower than what might be suggested by the total frame count.
While the Waymo Open Dataset \cite{sun2020scalability} is much larger in comparison (with 230,000 annotated frames), the dataset has a restrictive license.

An additional challenge with lidar datasets is that—unlike image datasets, where models generally transfer well when applied to images captured on different cameras—point cloud models perform poorly when applied to data captured with different sensors \cite{yi2021complete, ryu2023instant}.
Lastly, importantly, \textit{all} of the aforementioned lidar datasets are focused on autonomous driving, so their utility in other sectors (e.g.,  agriculture, construction, and mining) is extremely limited.
A method making it possible to acquire large, labeled lidar datasets in a cost-effective way would thus be valuable in many industries.

To address the limited sample sizes and sample diversity in existing lidar datasets, a number of techniques incorporating data augmentations and/or computer-generated data have been proposed.
Many of the data augmentation approaches involve some variation of ``mixing'' scenes, whether that be ``cutting'' objects from one scene and ``pasting'' them into another scene \cite{yan2018second, lang2019pointpillars, ren2022object, hu2023context}, or combining entire scenes \cite{nekrasov2021mix3d, ren2022object}.
While these approaches can produce novel \textit{combinations} of input signals, the fundamental \textit{diversity} of the input signals (e.g., the number of distinct street scenes in an autonomous driving dataset) is still limited in small datasets.
To make this point clear, consider the Waymo Open Dataset, which features data from San Francisco, Phoenix, and Mountain View.
No amount of mixing in this dataset would allow a model to encounter a peacock crossing the road during training.\footnote{A common occurrence on the streets of Miami: \url{https://www.nbcmiami.com/news/local/peacocks-in-miami-should-they-stay-or-should-they-go/2702779/}.}
Likewise, the diversity and quality of computer-generated scenes are a function of the 3D assets and graphics engines made available through the rendering pipelines (e.g., \cite{dosovitskiy2017carla, hurl2019precise, fang2020augmented, xiao2022transfer, saltori2022cosmix, li2023pcgen, yang2023reconstructing}), which makes them unsuitable for many applications.
Importantly, \textit{all of these approaches rely on previously labeled data}, whether those labels are provided by human annotators or are automatically available via a computer-generated environment.

In this paper, we introduce Paved2Paradise (\textbf{P2P}; Figure \ref{fig:overview}), a simple, cost-effective approach for assembling large, diverse, realistic, and fully annotated lidar datasets \textit{from scratch}—all while requiring minimal human annotation.
Our key insight is that, by deliberately collecting \textit{separate} ``background'' and ``object'' datasets (i.e., ``factoring the real world''), we can intelligently combine them to produce a combinatorially large and diverse training set.
To demonstrate the efficacy of our approach, we used \textbf{P2P} to generate a dataset for a challenging task in which public data does not exist: human detection in orchards.
A PointPillars \cite{lang2019pointpillars} model trained exclusively on the \textbf{P2P} synthetic dataset is highly effective at detecting humans in orchards, including when individuals are occluded by tree branches.
We further performed a quantitative analysis of \textbf{P2P} by generating a pedestrian detection dataset that sources pedestrian-less backgrounds from KITTI, and we find that a \textbf{P2P}-trained model performs comparably to a model trained on the full KITTI dataset.

\section{Comparison to LiDARsim}

LiDARsim \cite{manivasagam2020lidarsim} is conceptually similar to \textbf{P2P} in motivation, but the two pipelines differ considerably in execution with \textbf{P2P} being both simpler and more flexible than LiDARsim.
LiDARsim builds a catalog of surfel \cite{pfister2000surfels} meshes consisting of ``static maps'' and ``dynamic objects'' that serve as assets for creating synthetic scenes as in a graphics engine.
To create a static map for a specific area, LiDARsim requires (on average) three lidar passes over the area.
The lidar scans are then fused with wheel-odometry, IMU, and GPS measurements and converted into a single merged point cloud using GraphSLAM \cite{thrun2006graph}, i.e., the merging step is rather complex and difficult to use.\footnote{The hdl\_graph\_slam ROS package (\url{https://github.com/koide3/hdl_graph_slam}) has 116 configurable parameters and notes that ``mapping quality largely depends on the parameter setting''.}
Finally, the merged point cloud is converted into a surfel mesh, which also involves parameter tuning to achieve optimal results.

For their dynamic objects, \citet{manivasagam2020lidarsim} focused exclusively on vehicles (non-rigid objects—like humans—were explicitly omitted), which is reflected in the LiDARsim asset creation process.
Specifically, to create a dynamic object, LiDARsim first accumulates all of the points found in an object's bounding box over a short time sequence and then mirrors the points over the object's heading (which exploits the bilateral symmetry commonly found in vehicles).
The point cloud is then refined with an iterative closest point algorithm \cite{park2017colored} (using lidar intensity as the color value) and finally converted into a surfel mesh.
Importantly, dynamic object creation in LiDARsim depends on lidar scenes that have \textit{already been annotated} with bounding boxes, which are unavailable in many applications and are costly and time-consuming to obtain.
Further, the diversity of behaviors reflected in the dynamic object dataset are left to chance, i.e., many behaviors are likely to be missing.

In contrast to LiDARsim, \textbf{P2P} leverages the raw lidar scans directly, which means: (1) new background data is immediately available for synthetic scene creation (i.e., without requiring additional preprocessing or access to auxiliary inputs) and (2) the approach works equally well with rigid and non-rigid objects.
Further, \textbf{P2P} only requires a single scan per static map, so \textbf{P2P} can collect at least three times as many static maps as LiDARsim for a given data collection time budget.
Lastly, because dynamic object data collection is a separate and deliberate process in \textbf{P2P}, the diversity of object behaviors is directly under the researcher's control instead of being left to serendipity.

\section{Paved2Paradise}
\vspace{-8pt}

\epigraph{\twoBeamedQuavers{} Don't it always seem to go\\
That the data that you want would take too long?\\
With Paved2Paradise \\
You only need people and a parking lot \twoBeamedQuavers{}}{``Little Red MULE'' - \textit{JonAI Mitchell}\footnotemark}
\vspace{-8pt}
\footnotetext{If you find these lyrics funny, they were written by the authors.
If not, they were written by ChatGPT.}

\noindent
To motivate the \textbf{P2P} pipeline, note that the ``ideal'' 3D object detection dataset (i.e., one that reflects the diversity of scenarios a model may experience in the wild) will vary across four factors: (1) \textbf{instances} of the object class(es), (2) \textbf{poses} each object can take on, (3) \textbf{locations} where an object can be found in a scene relative to a lidar sensor, and (4) environmental \textbf{backgrounds} where the objects can be seen.
Clearly, directly acquiring and annotating a dataset that fully reflects this combinatorial diversity is impractical.
Our key insights are that: (1) \textbf{instance} and \textbf{background} datasets can be collected \textit{separately} with the \textbf{background} dataset collected passively, (2) the \textbf{location} factor can be entirely simulated, which means (3) \textbf{poses} only need to be collected from a single location with respect to the lidar sensor.
The full \textbf{P2P} pipeline thus consists of the following four steps:

\begin{enumerate}
    \item Collecting copious background data that reflects the variety of environmental conditions a model will encounter in the wild (Section \ref{sec:background}).
    \item Recording individuals from the desired object class(es) performing different behaviors at an isolated, flat location (Section \ref{sec:object}).
    \item Bootstrapping labels for the object dataset (Section \ref{sec:bootstrap}).
    \item Generating training samples by combining randomly sampled objects with randomly sampled backgrounds in such a way that the synthetic scenes are consistent with the sensor's perspective (Sections \ref{sec:level}, \ref{sec:place}, \ref{sec:combine}, and \ref{sec:occlude}).
\end{enumerate}

\subsection{Collecting the background scenes}\label{sec:background}

The goal of the background collection step of \textbf{P2P} is to capture as much environmental data as possible.
The only requirement for this step is that the collected scenes do \textit{not} contain any objects that the model will be trained to detect.
Because these environmental captures will serve as the backgrounds into which objects will be placed, any scenes that contain \textit{unlabeled} objects are in fact incorrectly labeled and will hurt model performance.

In many cases, the backgrounds can be collected passively because (presumably) the task that needs to be automated has not yet been automated.
As a result, data can be recorded during standard vehicle/equipment operation to produce a large and diverse background dataset that spans environmental conditions—the operator simply needs to diligently note moments when an object from a detection category is observed so that the surrounding frames can be removed from the background pool.
In the case of pedestrian detection for autonomous driving, background data could be collected by driving in neighborhoods when they experience their lowest pedestrian traffic.

\subsection{Collecting the object scenes}\label{sec:object}

Note that, if given a mesh of a 3D object, simulating the point cloud for that object from the perspective of a lidar sensor is trivial using ray casting (ignoring occlusion and sensor effects for the time being).
Indeed, this is an effective approach as demonstrated by LiDAR-Aug \cite{fang2021lidar} and LiDARsim \cite{manivasagam2020lidarsim}, and can be taken to the extreme by assembling datasets that are \textit{entirely} computer-generated (e.g., \cite{dosovitskiy2017carla, hurl2019precise, xiao2022transfer, saltori2022cosmix}).
But, for non-rigid objects, simple meshes are insufficient because their real-life analogs undergo both rigid \textit{and} non-rigid transformations.
Using high-quality ``rigged'' meshes that can be animated is one potential solution to the non-rigidity issue.
However, such meshes are generally expensive to acquire.
Further, animating each separate mesh (of which there should be many to reflect the natural diversity of the object class) requires employing/contracting 3D animators (although deep learning may be able to automate animation in the future \cite{mourot2022survey}).

The goal of the object collection step in \textbf{P2P} is to match the flexibility and diversity of this hypothetical animation library in a cost-effective and time-efficient way.
To motivate our approach, we first describe a concrete example of a naive strategy for collecting the object dataset.
Imagine that a researcher would like to be able to detect humans in a $200 \times 200$ square grid located directly in front of a lidar sensor.
Further, assume that these particular humans come in $200$ different body types and can take on one of $200$ different poses.
There are thus $200^{4} = 1.6$ \textit{billion} unique human point clouds (in terms of the point coordinates) that a model could encounter for this particular task (again, ignoring any sensor effects).

In theory, recording an individual of each body type performing each pose in each grid cell at the collection location would provide the researcher with a full object dataset that, following annotation, could be mixed with background data.
Of course, collecting and annotating 1.6 billion samples is prohibitively costly and/or time-consuming.
\textbf{P2P} uses a simpler, far less time-intensive approach—recording individuals performing a variety of natural behaviors at a \textit{single} location relative to the sensor\footnote{Specifically, as close to the sensor as possible while maintaining full visibility of the human.} in an isolated, flat setting (such as an empty area of a parking lot).
This strategy immediately reduces the data collection workload by a factor of 40,000, i.e., only 40,000 samples (the scale of the SemanticKITTI dataset) would need to be labeled in this hypothetical example.

\subsection{Bootstrapping labels for the object scenes}\label{sec:bootstrap}

Because the \textbf{P2P} object scenes are so simple and consistent, a PointNet++ \cite{pointnetplusplus} regression model can be trained to accurately predict bounding box attributes from a relatively small number of human-labeled samples.
Specifically, a sample is created by (1) extracting only those points in the performance area that are \textit{above the ground plane} and (2) centering the extracted points based on the median coordinate values.
Once the PointNet++ model reaches an acceptable level of accuracy, the remaining samples can be automatically labeled.

\subsection{Leveling the scenes}\label{sec:level}

At this point, \textbf{P2P} has a collection of $M$ background point clouds $\mathcal{B} = \{\mathbf{B}_{m}\}_{m=1}^{M}$ where $\mathbf{B}_{m} = \{\mathbf{b}_{m,p}\}_{p=1}^{P_{m}}$ is a set of $P_{m}$ points with $\mathbf{b}_{m,p} = (x_{\mathbf{b}_{m,p}}, y_{\mathbf{b}_{m,p}}, z_{\mathbf{b}_{m,p}})$.
Likewise, \textbf{P2P} has a collection of $N$ object \textit{scene}\footnote{Note that we are explicitly distinguishing an object \textit{scene} point cloud from an the isolated object point cloud.} point clouds $\mathcal{O} = \{\mathbf{O}_{n}\}_{n=1}^{N}$ where $\mathbf{O}_{n} = \{\mathbf{o}_{n,q}\}_{n=1}^{Q_{n}}$ is a set of $Q_{n}$ points with $\mathbf{o}_{n,q} = (x_{\mathbf{o}_{n,q}}, y_{\mathbf{o}_{n,q}}, z_{\mathbf{o}_{n,q}})$.
Further, each object scene has an associated bounding box for the object defined by its center $\mathbf{c}_{n} = (x_{\mathbf{c}_{n}}, y_{\mathbf{c}_{n}}, z_{\mathbf{c}_{n}})$, extent $\mathbf{d}_{n} = (x_{\mathbf{d}_{n}}, y_{\mathbf{d}_{n}}, z_{\mathbf{d}_{n}})$, and rotation matrix $\mathbf{R}_{n}$.
For both collections of point clouds, the points are in the sensor's coordinate frame, i.e., the sensor is located at the origin with the $x$-axis pointing forward out of the sensor, the $y$-axis pointing to the sensor's left, and the $z$-axis pointing up.

\begin{figure}[htbp]
\centering
\includegraphics[scale=0.21]{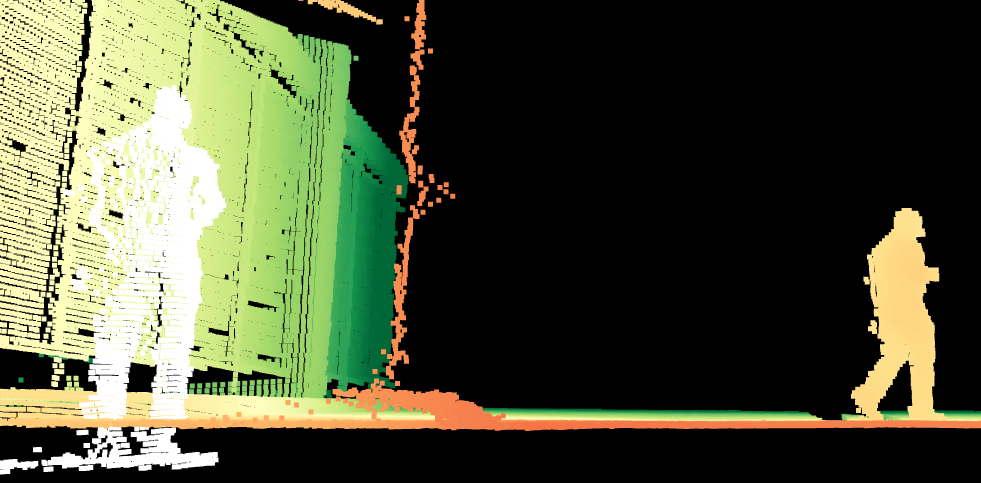}
\vspace{-4pt}
\setlength{\belowcaptionskip}{-16pt}
\caption{Because this sensor's $xy$-plane was not level with the ground plane, naively rotating the human point cloud around the sensor's $z$-axis results in the human's feet being underground.}
\label{fig:naive_rotation}
\end{figure}

\begin{figure}[htbp]
\centering
\vspace{-4pt}
\includegraphics[scale=0.44]{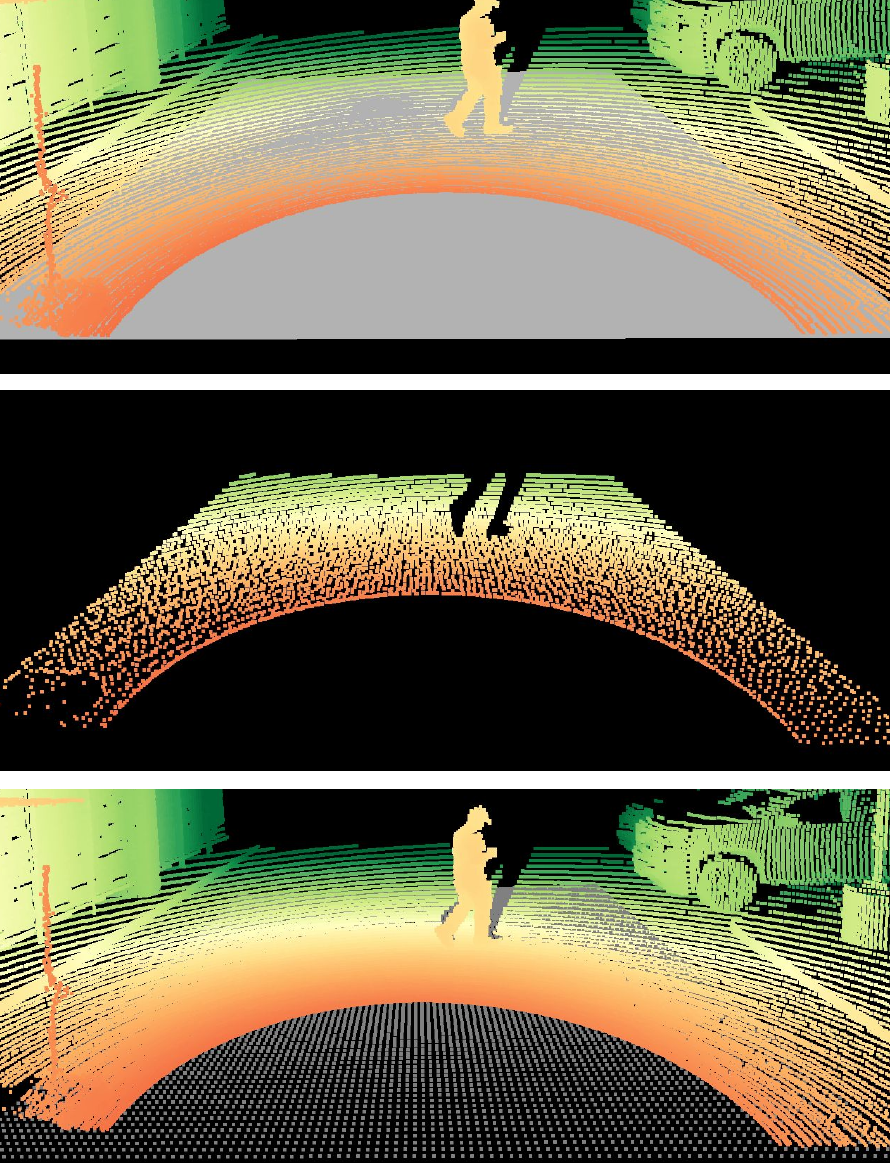}
\setlength{\belowcaptionskip}{-8pt}
\caption{The Paved2Paradise leveling procedure estimates the ground plane for a scene (top) by performing linear regression on ground points (middle) obtained by finding the nearest neighbors to a set of grid points beneath the scene (bottom).}
\label{fig:level}
\end{figure}

Because the sensors used to obtain the background and object scenes may not be perfectly level with their respective ground planes nor the same heights off the ground, naively manipulating and combining the point clouds can produce unrealistic scenes (Figure \ref{fig:naive_rotation}).
Therefore, it is necessary to ``level'' the point clouds before proceeding (Figure \ref{fig:level}).
The leveling procedure begins by setting up a grid of $G \times G$ points in a predetermined region $x \in [x_{\text{min}}, x_{\text{max}}]$ and $y \in [-y_{\text{max}}, y_{\text{max}}]$ where $x_{\text{min}}$, $x_{\text{max}}$, and $y_{\text{max}}$ are provided by the user, and the $z$ value for each grid point is the minimum $z$ value of the scene points in the region.
Next, the ``ground points'' are defined as the scene points in the grid region that are nearest neighbors with the grid points.

A linear regression model of the form $z = b_{0} + b_{1} x + b_{2} y$ is then fit to the ground points to estimate the ground plane.
The normal vector $\textbf{h}_{\mathds{1}}$ for the estimated ground plane is thus $\textbf{h}_{\mathds{1}} = \frac{\mathbf{h}}{\lVert \mathbf{h} \rVert}$ where $\textbf{h} = [-b_{1}, -b_{2}, 1]$.
The rotation matrix $\mathbf{R}_{\{\mathbf{O},\mathbf{B}\}}$ for leveling the estimated ground plane can be obtained using Rodrigues' rotation formula, i.e., $\mathbf{R}_{\{\mathbf{O},\mathbf{B}\}} = \mathbf{I} + [\mathbf{v}]_{\times} + [\mathbf{v}]_{\times}^2\frac{1}{1 + \textbf{h}[3]}$ where:

\[
[\mathbf{v}]_{\times} \stackrel{\rm def}{=} \begin{bmatrix}
\,\,0 & \!-v_3 & \,\,\,v_2\\
\,\,\,v_3 & 0 & \!-v_1\\
\!-v_2 & \,\,v_1 &\,\,0
\end{bmatrix}
\]

\noindent
and $\mathbf{v} = \mathbf{h}_{\mathds{1}} \times [0, 0, 1]$.\footnote{See: \url{https://en.wikipedia.org/wiki/Rodrigues\%27_rotation_formula\#Matrix_notation} and: \url{https://math.stackexchange.com/a/476311/614328}.}
\textbf{P2P} thus levels an object scene point cloud $\mathbf{O}$\footnote{We omit the subscripts here for succinctness.} such that its estimated ground plane is aligned with the $xy$-plane at $z = 0$ using the following transformation:

\vspace{-0.2cm}
\begin{equation}
    \widetilde{\mathbf{O}} = \mathbf{R}_{\mathbf{O}} \mathbf{O} - \mathbf{t}_{\mathbf{O}}
\end{equation}

\noindent
where $\mathbf{R}_{\mathbf{O}}$ is the leveling rotation matrix for the object scene and $\mathbf{t}_{\mathbf{O}} = [0, 0, b_{0}]$.
Using the same procedure, \textbf{P2P} can convert a background point cloud $\mathbf{B}$ into a leveled background $\widetilde{\mathbf{B}}$.
\subsection{Placing an object}\label{sec:place}

While a large dataset can be assembled by combining unmodified object point clouds with background point clouds, the true diversity of the dataset will be relatively low because the object point clouds will always be found in the same locations.
If an object point cloud could instead be placed at an \textit{arbitrary} location in a background point cloud, the diversity of the synthetic dataset would dramatically increase.
However, due to the sensor's perspective of the scene, placing an object point cloud at an arbitrary location in a background point cloud is not as simple as translating the object point cloud to the new location.

\begin{figure}[htbp]
\centering
\includegraphics[scale=0.45]{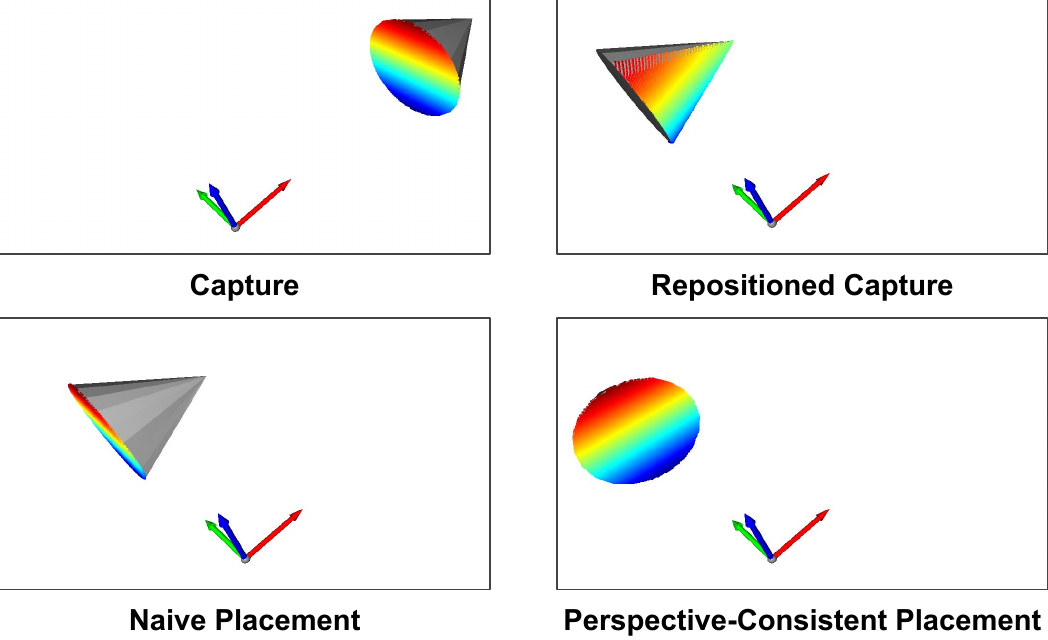}
\setlength{\belowcaptionskip}{-16pt}
\caption{Naively placing a point cloud at an arbitrary location in a scene can produce a synthetic lidar scan that is physically impossible.
\textbf{Top Left}: In the original scene, the cone is directly in front of the sensor (indicated by the axes) with the cone's base perpendicular to the sensor's $x$-axis.
\textbf{Top Right}: Physically moving the cone to the left of the sensor (without rotating the object) results in an entirely different lidar scan.
\textbf{Bottom Left}: However, naively translating the point cloud from the original scene to the left of the sensor leads to an impossible lidar scan.
\textbf{Bottom Right}: Rotating the translated point cloud produces a perspective-consistent synthetic lidar scan.
}
\vspace{-0.2cm}
\label{fig:perspective}
\end{figure}

For example, consider a scene where a cone has been placed directly in front of the sensor with the base oriented perpendicular to the sensor's $x$-axis (Figure \ref{fig:perspective}).
In this case, the sensor will see a disk of points.
If the cone was then physically placed—without rotating the object—directly to the left of the sensor, the sensor would see a curved triangle of points.
However, naively translating the disk of points from the original scene to the side of the sensor would result in an impossible point cloud from the sensor's perspective.
Not only is the side of the cone missing, but a side view of a disk should produce a line of points, i.e., the new point cloud is wrong in two different ways. 

A straightforward way to enforce perspective-consistency is by restricting object movements to (1) translations along the direction defined by the object's center and (2) rotations around the $z$-axis of the sensor.
Therefore, given an arbitrary location on the leveled ground plane $\mathbf{\rho} = (x_{\mathbf{\rho}}, y_{\mathbf{\rho}}, 0)$, the extracted and repositioned object point cloud $\widetilde{\mathbf{O}}_{\rho}$ is obtained with:

\vspace{-0.2cm}
\begin{equation}\label{eq:place}
    \widetilde{\mathbf{O}}_{\rho} = \mathbf{R}_{\rho}(\psi(\widetilde{\mathbf{O}}, \widetilde{\mathbf{c}}, \widetilde{\mathbf{d}}, \widetilde{\mathbf{R}}) + \mathbf{t}_{\mathbf{\rho}})
\end{equation}

\noindent
where $\mathbf{R}_{\rho} = \big(\begin{smallmatrix}
  \cos(\theta) & -\sin(\theta) & 0\\
  \sin(\theta) & \cos(\theta) & 0\\
  0 & 0 & 1
\end{smallmatrix}\big)$, $\theta = \text{atan2}(y_{\mathbf{\rho}}, x_{\mathbf{\rho}}) - \text{atan2}(y_{\tilde{\mathbf{c}}}, x_{\tilde{\mathbf{c}}})$, $\psi$ is a cropping function, $\widetilde{\mathbf{c}}$, $\widetilde{\mathbf{d}}$, and $\widetilde{\mathbf{R}}$ are the leveled bounding box parameters,  $\mathbf{t}_{\rho} = \lVert \mathbf{\rho} \rVert \frac{\tilde{\mathbf{c}}_{0}}{\lVert \tilde{\mathbf{c}}_{0} \rVert} - \tilde{\mathbf{c}}_{0}$, and $\tilde{\mathbf{c}}_{0} = (x_{\tilde{\mathbf{c}}}, y_{\tilde{\mathbf{c}}}, 0)$.

\subsection{Combining an object with a background}\label{sec:combine}

At this point, the repositioned object can be placed into a leveled background by combining the two sets of points, i.e., $\widetilde{\mathbf{S}} = \widetilde{\mathbf{B}} \cup \widetilde{\mathbf{O}}_{\rho}$.
The leveled synthetic composite scene $\widetilde{\mathbf{S}}$ can then be transformed back into the original background coordinate frame with $\mathbf{R}_{\mathbf{B}}^{\top} (\widetilde{\mathbf{S}} + \mathbf{t}_{\mathbf{B}}) = \mathbf{S}_{u}$.
While these steps conceptually describe how \textbf{P2P} combines an object with a background, they are inefficient because the background points are simply returned to their original positions.
As a result, in practice, the background inverse transformation is only applied to $\widetilde{\mathbf{O}}_{\rho}$, i.e.:

\vspace{-0.2cm}
\begin{equation}\label{eq:unlevel}
    \widehat{\mathbf{O}} = \mathbf{R}_{\mathbf{B}}^{\top} (\widetilde{\mathbf{O}}_{\rho} + \mathbf{t}_{\mathbf{B}})
\end{equation}

\noindent
The points in $\widehat{\mathbf{O}}$ are then re-indexed such that $i = \text{rank}(q, \hat{\mathbf{q}})$ where 
$\hat{\mathbf{q}} = \{q \in \{1 \dots Q\} : \hat{\mathbf{o}}_{q} \in \widehat{\mathbf{O}} \}$ and $\text{rank}(q, \hat{\mathbf{q}})$ returns the ranking of $q$ in the set of indexes $\hat{\mathbf{q}}$.
The unoccluded composite scene $\mathbf{S}_{u}$ is thus:

\vspace{-0.25cm}
\begin{equation}
    \mathbf{S}_{u} = \mathbf{B} \cup \widehat{\mathbf{O}}
\end{equation}

\subsection{Simulating the final LiDAR scan}\label{sec:occlude}

To make the final point cloud appear more realistic, \textbf{P2P} simulates occlusion and sensor effects.
To simulate occlusion, \textbf{P2P} uses an \textit{approximate} ray intersection algorithm (similar to \citet{schaufler2000ray}), which occludes points in both the object point cloud and the background point cloud to create an occluded point cloud scene.
Because an object is only occluded when lidar beams in its sector are blocked, the occlusion algorithm begins by extracting a subset of background points $\mathbf{B}_{\alpha} \subset \mathbf{B}$ that are in the same sector as $\widehat{\mathbf{O}}$, i.e.:

\vspace{-0.4cm}
\[
    \mathbf{B}_{\alpha} = \{ \mathbf{b}_{p} \in \mathbf{B} : \min(\alpha(\widehat{\mathbf{O}})) - \epsilon < \alpha(\mathbf{b}_{p}) < \max(\alpha(\widehat{\mathbf{O}})) + \epsilon\}
\]

\noindent
where $\alpha$ is a function that returns the azimuth of a point, $\alpha(\widehat{\mathbf{O}}) = \{\alpha(\hat{\mathbf{o}}_{i}) \}_{i=1}^{|\widehat{\mathbf{O}}|}$, and $\epsilon$ is a user-defined tolerance factor.
Further, because an object can only be occluded by a second object if the second object is \textit{closer} to the sensor than the object, the occlusion algorithm extracts two additional subsets of points from $\mathbf{B}_{\alpha}$:

\vspace{-0.4cm}
\begin{align*}
    \mathbf{B}_{\beta} &= \{ \mathbf{b}_{p} \in \mathbf{B}_{\alpha} : \lVert \mathbf{b}_{p} \rVert \leq \max(\lVert \widehat{\mathbf{O}} \rVert) \} \\
    \mathbf{B}_{\gamma} &= \{ \mathbf{b}_{p} \in \mathbf{B}_{\alpha} : \lVert \mathbf{b}_{p} \rVert \geq \min(\lVert \widehat{\mathbf{O}} \rVert) \}
\end{align*}

\noindent
where $\lVert \widehat{\mathbf{O}} \rVert = \{ \lVert \hat{\mathbf{o}}_{i} \rVert \}_{i=1}^{|\widehat{\mathbf{O}}|}$.
$\mathbf{B}_{\beta}$ will be used to occlude points in $\widehat{\mathbf{O}}$, while $\widehat{\mathbf{O}}$ will be used to occlude points in $\mathbf{B}_{\gamma}$.

\begin{figure}[htbp]
\centering
\includegraphics[scale=1.6]{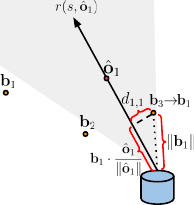}
\vspace{-0.1cm}
\caption{The Paved2Paradise object occluding procedure.
Given a set of object points $\widehat{\mathbf{O}}$ (here, $\widehat{\mathbf{O}}$ contains a single point, i.e., $\widehat{\mathbf{O}} = \{\hat{\mathbf{o}}_{1}\}$), Paved2Paradise extracts a subset of (reindexed; see Section \ref{sec:occlude}) background points $\mathbf{B}_{\beta} \subset \mathbf{B}$ such that each $\mathbf{b}_{j} \in \mathbf{B}_{\beta}$ is \textit{both} in the same sector as $\widehat{\mathbf{O}}$ (delineated by the gray cone) \textit{and} closer to the sensor (the blue cylinder) than at least one point in $\widehat{\mathbf{O}}$.
Next, for each object point $\hat{\mathbf{o}}_{i}$, Paved2Paradise calculates the minimum distances $d_{i,j}$ between the points in $\mathbf{B}_{\beta}$ and the ray $r(s, \hat{\mathbf{o}}_{i}) = s \frac{\hat{\mathbf{o}}_{i}}{\lVert \hat{\mathbf{o}}_{i} \rVert}$.
Because the closest point along $r(s, \hat{\mathbf{o}}_{i})$ to $\mathbf{b}_{j}$ is the projection of $\mathbf{b}_{j}$ onto the ray, i.e., $(\mathbf{b}_{j} \cdot \frac{\hat{\mathbf{o}}_{i}}{\lVert \hat{\mathbf{o}}_{i} \rVert}) \frac{\hat{\mathbf{o}}_{i}}{\lVert \hat{\mathbf{o}}_{i} \rVert}$, each $d_{i,j}$ can be calculated using the Pythagorean theorem.
Finally, if any of the $d_{i,j}$ values are less than some user-defined threshold, then $\hat{\mathbf{o}}_{i}$ is considered ``occluded'' and dropped from the final point cloud for the scene.
}
\label{fig:occlusion}
\vspace{-0.2cm}
\end{figure}

We now describe the \textbf{P2P} approximate ray intersection function $\phi$, which has a signature of $\mathbf{D}_{\phi} = \phi(\mathbf{D}, \mathbf{E}, F)$ where $\mathbf{E}$ is a point cloud that will be used to occlude points in another point cloud $\mathbf{D}$ and $F$ is a user-defined threshold.
Here, we provide details for the case of $\widehat{\mathbf{O}}_{\phi} = \phi(\widehat{\mathbf{O}}, \mathbf{B}_{\beta}, F_{o})$ (Figure \ref{fig:occlusion}), but the procedure is identical for $\mathbf{B}_{\phi} = \phi(\mathbf{B}_{\gamma}, \widehat{\mathbf{O}}, F_{b})$.
First, the points in $\mathbf{B}_{\beta}$ are re-indexed such that $j = \text{rank}(p, \mathbf{p})$ where 
$\mathbf{p} = \{p \in \{1 \dots P\} : \mathbf{b}_{p} \in \mathbf{B}_{\beta} \}$.
Next, for each object point $\hat{\mathbf{o}}_{i}$, the procedure calculates the minimum distances $d_{i,j}$ between the points in $\mathbf{B}_{\beta}$ and the ray $r(s, \hat{\mathbf{o}}_{i}) = s \frac{\hat{\mathbf{o}}_{i}}{\lVert \hat{\mathbf{o}}_{i} \rVert}$ where $s \in (0, \infty)$.
Because the closest location on a ray to a given point is the projection of that point onto the ray, each $d_{i,j}$ can be calculated using the Pythagorean theorem, i.e.:

\vspace{-0.2cm}
\begin{equation}
    d_{i,j} = \sqrt{\lVert \mathbf{b}_{j} \rVert^{2} - (\mathbf{b}_{j} \cdot \frac{\hat{\mathbf{o}}_{i}}{\lVert \hat{\mathbf{o}}_{i} \rVert})^{2}}
\end{equation}

\noindent
If $\min(\{d_{i,j}\}_{j=1}^{J}) < F_{o}$, $\hat{\mathbf{o}}_{i}$ is considered ``occluded'' and dropped from the final scene.

\begin{figure}[htbp]
\centering
\includegraphics[scale=1.3]{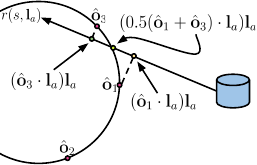}
\setlength{\belowcaptionskip}{-8pt}
\caption{To simulate the appearance of an object's point cloud at different distances relative to the sensor, Paved2Paradise uses an ``unoccluding'' procedure that is essentially the reverse of the occluding approach shown in Figure \ref{fig:occlusion}.
Specifically, an object's simulated surface point for a specific lidar beam $ \mathbf{l}_{a}$ is calculated by taking the average of the projected object points $(\hat{\mathbf{o}}_{i} \cdot \mathbf{l}_{a}) \mathbf{l}_{a}$ on the ray $r(s, \mathbf{l}_{a})$ for the two closest object points.
}
\label{fig:unocclude}
% \vspace{-0.1cm}
\end{figure}

With regards to sensor effects, note that the number and locations of points detected on an object is a function of its position relative to a sensor.
For example, when an object is moved closer to a sensor, more lidar beams strike the object, which produces a point cloud with more points.
To simulate the appearance of the object point cloud at different distances, \textbf{P2P} uses an ``unoccluding'' function $\zeta$ (Figure \ref{fig:unocclude}) that in some sense is the reverse of  the \textbf{P2P} occluding procedure.

Specifically, given a ray $r(s, \mathbf{l}_{a}) = s \mathbf{l}_{a}$ where $\mathbf{l}_{a}$ is the direction (i.e., a unit vector) corresponding to the lidar beam indexed by $a$, the procedure calculates the distances $d_{a,i}$ to each object point $\hat{\mathbf{o}}_{i}$ using the same approach described for the occluding function.
If no object points have a distance less than some user-defined threshold $L$ to the ray, then the ray does not intersect with the object and no surface point is generated for that lidar beam.
In the case where at least one $d_{a,i} < L$, the simplest strategy would be to take the point on the ray $(\hat{\mathbf{o}}_{i} \cdot \mathbf{l}_{a}) \mathbf{l}_{a}$ corresponding to the minimum $d_{a,i}$ as the surface point.
However, this point is often relatively far from the lidar beam's true intersection point, which causes the simulated object point cloud to appear ``jagged''.
Instead, the \textbf{P2P} procedure takes \textit{up to} two of the closest object points to the ray.
If only one point meets this criterion, then that point is used as the object's surface point if $d_{a,i} < \frac{L}{2}$ .
If there are two close points, the surface point is their average.
The final simulated scene $\mathbf{S}$ is thus:\footnote{Note that multiple objects can be incorporated into a scene by following a two step algorithm: (1) apply \textbf{P2P} to a sampled object and background, (2) sample a new object and apply \textbf{P2P} while treating the previously created P2P scene as the background.}

\vspace{-0.3cm}
\begin{equation}
    \mathbf{S} = (\mathbf{B} \setminus \mathbf{B}_{\alpha}) \cup (\mathbf{B}_{\alpha} \setminus \mathbf{B}_{\gamma}) \cup \mathbf{B}_{\phi} \cup \widehat{\mathbf{O}}_{\zeta}
\end{equation}

A last practical consideration is that of storage.
For our \textbf{orchard} dataset (Section \ref{sec:experiments}), each synthetic scene contained $\sim$70,000 points, which would require $\sim$850 kB of storage per sample if each scene was saved in the PCD format.
Therefore, our 500,000 synthetic scenes would require $\sim$425 GB of storage if each point cloud was stored in full.
Such a storage rate introduces inconvenient dataset accessibility challenges (e.g., requiring datasets to be downloaded from a remote server whenever they are needed). 
Fortunately, the storage footprint for a \textbf{P2P} synthetic dataset can be considerably reduced by saving only: (1) the final transformed and occluded object point clouds and (2) the \textit{indexes} of the occluded background points.
Because the number of object points and occluded background points is much smaller than the total number of points in the final scene, the storage requirements are greatly reduced.
During training, the synthetic scene for a given sample is assembled on the fly by: (1) loading the (transformed) object and (original) background point clouds, (2) selecting the \textit{unoccluded} background points based on the occluded indexes, and (3) combining the two point clouds.

\section{Experiments}\label{sec:experiments}

To demonstrate the utility of \textbf{P2P}, we used the pipeline to generate synthetic datasets for two tasks: (1) human detection in orchards and (2) pedestrian detection in urban environments.
We refer to these tasks as the \textbf{orchard} and \textbf{urban} tasks.
The \textbf{orchard} task is an excellent case study for \textbf{P2P} because public datasets of humans in orchards do not exist.
Further, the task is inherently challenging due to: (i) the wide variety of poses individuals working in an orchard can be found in and (ii) the frequency with which individuals are partially occluded by trees.

For the \textbf{orchard} task, we collected 6,717 proprietary background scenes by driving a Kawasaki MULE equipped with an Ouster OS1 (REV7) lidar sensor for $\sim$20 minutes through a walnut orchard located in Central California.
For the \textbf{urban} task, we sourced 5,581 background scenes from the KITTI training dataset.
Specifically, we randomly split the 7,481 KITTI training frames 90/10 into training and test sets consisting of 6,732 and 749 frames, which we refer to as the \textbf{urban} training and test sets.
We then used all of the frames from the \textbf{urban} training set that did not have any annotated \texttt{Pedestrian} bounding boxes in our detection region as background scenes.

For both the \textbf{orchard} and \textbf{urban} tasks, we sourced objects from the same proprietary set of 21,263 object scenes, but we excluded crouching poses from the \textbf{urban} synthetic dataset leaving 11,054 object scenes for that task.
The object scenes were collected by recording 48 different individuals performing a pose protocol in a parking lot over the span of eight hours.
The pose protocol began with the individual standing on a flat rubber dot located 5 m in front of a lidar sensor.
The individual was then instructed to slowly perform one full rotation (either clockwise or counterclockwise as determined by coin flip) in place while maintaining a neutral stance.
Next, the individual was instructed to stand at a randomly selected location on a 1 m radius chalk circle surrounding the dot and then walk in a full circle in the clockwise direction.
The individual was then instructed to stand at a different randomly selected location on the chalk circle and walk in a full circle in the \textit{counterclockwise} direction.
Finally, the previous two steps were repeated, but with the individual receiving the additional instruction to pantomime walking through trees (e.g., crouching and/or moving their arms as if they were moving branches out of the way).
Once the human data was collected, we annotated a subset of the samples with bounding boxes using Segments.ai.
We then trained a PointNet++ model on the annotated subset and used the trained model to generate the bounding boxes for the remaining samples (see Section \ref{sec:pointnet} in the supplement for training details).

For each task, we generated 500,000 synthetic samples that were split 99.5/0.5 into training and validation sets of 497,500 and 2,500 samples.
For the \textbf{orchard} task, the detection region bounds were $(0, 12)$, $(-4.625, 4.625)$, and $(-1, 5)$ for the $x$, $y$, and $z$ dimensions, and the simulated lidar consisted of beams at 128 evenly spaced elevation angles from -22.5° to 22.5° and 2,048 evenly spaced azimuth angles from 0° to 360°.
For the \textbf{urban} task, the detection region bounds were $(0, 19)$, $(-9, 9)$, and $(-2.5, 4)$ for the $x$, $y$, and $z$ dimensions, and the simulated lidar consisted of beams at 64 evenly spaced elevation angles from -24.8° to 2° and 2,083 evenly spaced azimuth angles from 0° to 360°.
For each task, we set $F_{b} = 3$ cm and $L = 4$ cm.
We set $F_{o} = 4$ cm and $F_{o} = 8$ cm for the \textbf{orchard} and \textbf{urban} tasks, respectively.
We doubled the number of background and object scenes for each task by mirroring each point cloud across the $x$-axis of the sensor.
We further augmented the object scenes by placing the human point clouds at randomly selected locations on the ground plane in the detection prism.
Lastly, for the \textbf{orchard} task, we only inserted a single object into each synthetic scene, while for the \textbf{urban} task we inserted up to 10 objects in a scene. 

For each task, we trained a neural network to perform a simplified object detection task—predicting the cell in a 2D grid that contains the center of the object's bounding box.
Specifically, the target $\mathbf{Y}$ is a 2D binary grid full of zeros except for the cells corresponding to bounding box centers, which contain ones.
This task is equivalent to pixel-wise binary classification and thus a model can be trained with the cross-entropy loss and evaluated threshold-free by calculating the average negative log-likelihood (NLL) and area under the precision-recall curve (AUC).
For the \textbf{orchard} task, we evaluated the model on a two minute recording of a drive through a walnut orchard that features two individuals acting naturally and coming in and out of view.
For the \textbf{urban} task, we evaluated both models on our \textbf{urban} test set.

Recently, \citet{li2023pillarnext} showed PointPillars models can match the performance of newer architectures with minor architectural and/or training modifications, so we used PointPillars as our neural network for all experiments.
Specifically, we trained three models: (1) a \textbf{P2P} \textbf{orchard} model, (2) a \textbf{P2P} \textbf{urban} model, and (3) a \textbf{baseline} \textbf{urban} model trained on our \textbf{urban} training set (which we further split 90/10 into training and validation sets consisting of 6,058 and 674 samples).
See Section \ref{sec:pointpillars} in the supplement for training details.

\vspace{-0.1cm}
\section{Results}\label{sec:results}
\vspace{-0.1cm}

\begin{figure}[htbp]
\centering
\vspace{-0.1cm}
% Frame 784.
\includegraphics[scale=0.18]{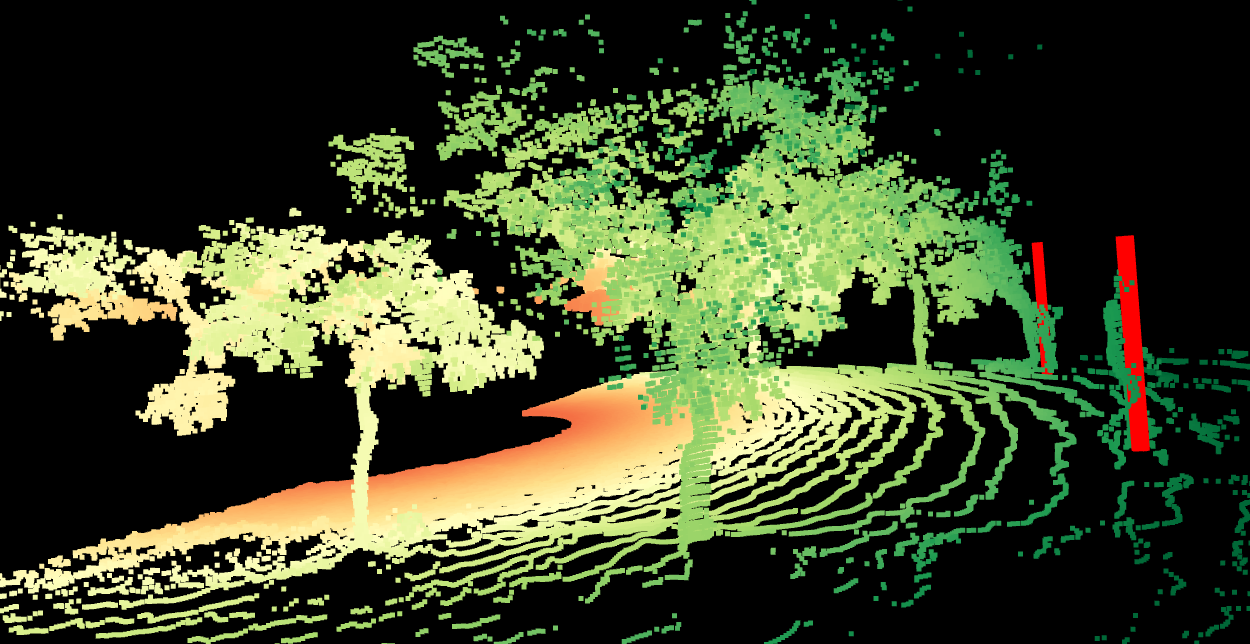}
\caption{A PointPillars model trained \textit{exclusively} on synthetic data generated by Paved2Paradise can reliably detect humans in orchards—including individuals who are heavily occluded by tree branches.
The vertical red bars indicate the model's estimates of the bounding box centers for the humans.}
\label{fig:orchard_detection}
\vspace{-0.2cm}
\end{figure}

When using a detection probability threshold of 0.23 on the \textbf{orchard} evaluation sequence, the \textbf{P2P} model obtained 100\% recall with only one false detection across 1,247 frames (741 of which included at least one human).
The \textbf{P2P} model consistently detected humans under challenging conditions, such as when individuals were heavily occluded by tree branches as seen in Figure \ref{fig:orchard_detection}.\footnote{A video of the full sequence can be viewed at: 
\url{https://drive.google.com/file/d/1iyyELpPYilfiQLS1UK_4fLTU-YTXtani/view?usp=sharing}
% $<$anonymized$>$
.}
Notably, the model effortlessly generalized to detecting two humans in scenes even though the synthetic samples only ever contained a single individual at a time.

For the \textbf{urban} task, we report the NLL and AUC-PR at the grid level and scene level for our \textbf{urban} test set in Table \ref{tab:metrics}.
For the scene-level metrics, we assigned a point cloud a label of one if it contained at least one pedestrian and zero otherwise, and we used the model's \textit{maximum} logit as its prediction.
Note that our \textbf{urban} task setup \textit{heavily favors} the \textbf{baseline} model because of how the datasets were created.
Specifically, because the KITTI training set was derived from \textit{sequences} of data, certain locations are repeated numerous times in the dataset, and even specific \textit{individuals} appear in multiple samples; i.e., the \textbf{baseline} model is effectively allowed to ``peek'' at pedestrian samples in the \textbf{urban} test set.\footnote{Among the 113 frames in the KITTI training set that have greater than five pedestrians in our detection region, only nine different locations are represented, with the top three locations being found in 30, 27, and 20 samples, respectively.}
Further, because the \textbf{P2P} pedestrian dataset was not labeled by the same individuals who labeled the KITTI dataset, the \textbf{P2P} labels may systematically differ from the KITTI labels in a way that favors the \textbf{baseline} model.
Despite these significant disadvantages, the \textbf{P2P} model performed admirably, matching the \textbf{baseline} model's NLL and AUC at the scene level.
Like the \textbf{orchard} \textbf{P2P} model, the \textbf{urban} \textbf{P2P} model can effectively detect pedestrians in challenging conditions, such as crowded scenes (see Figure \ref{fig:crowd} in the supplement).

\begin{table}[htbp]
\vspace{-0.2cm}
\caption{For our pedestrian detection task, a PointPillars model trained \textit{exclusively} on Paved2Paradise synthetic data (P2P) matched the performance—as measured by the average negative log-likelihood (NLL) and area under the precision-recall curve (AUC)—of a model trained on the full KITTI dataset (Base) at the scene level (see Section \ref{sec:results}).}
\vspace{-0.1cm}
\centering
\begin{tabular}{@{}lllllll@{}}
\toprule
        & \multicolumn{2}{c}{Grid} & \multicolumn{2}{c}{Scene} \\ \midrule
Dataset & NLL         & AUC        & NLL         & AUC              \\ \midrule
Base    & 1.11        & 0.34       & 0.21        & 0.96             \\
P2P     & 1.52        & 0.23       & 0.21        & 0.95             \\ \bottomrule
\end{tabular}
\label{tab:metrics}
\vspace{-0.45cm}
\end{table}

\section{Conclusion}
\vspace{-0.1cm}

Our results show \textbf{P2P} is a cost-effective and scalable pipeline for generating diverse and realistic synthetic lidar datasets.
As a result, \textbf{P2P} could help accelerate point cloud model development in sectors where acquiring lidar datasets has previously been cost-prohibitive.
While the minimalist lidar simulator we used here was effective, a more sophisticated simulator that incorporates (potentially learned) sensor physics \cite{caccia2019deep, manivasagam2020lidarsim, nakashima2021learning, guillard2022learning, zyrianov2022learning, li2023pcgen, yang2023unisim} and/or weather effects \cite{hahner2021fog, kilic2021lidar, hahner2022lidar} could produce even more realistic scenes.

\section{Author Contributions}
MAA conceived and implemented the pipeline, designed and ran the experiments, and wrote the manuscript.
NS led data collection efforts, discussed research ideas, and facilitated publication.

\section{Acknowledgments}
We would like to thank Aaron Avila, Drew Avila, Samir Chowdhury, Will Nitsch, and Joey Passarello for collecting the parking lot and orchard data.
We would like to further thank Samir Chowdhury for his detailed feedback on the manuscript.
Lastly, we would like to thank all of the volunteers who participated in the parking lot data collection.

{\small
\bibliographystyle{ieeenat_fullname}
\bibliography{main}
}

\clearpage
\setcounter{page}{1}
\maketitlesupplementary

\section{PointNet++ Training Details}
\label{sec:pointnet}

For our PointNet++ model, we used the same Set Abstraction levels described in \citet{pointnetplusplus} for their semantic scene labeling network.
Max-pooling was applied to the output of the final Set Abstraction level, and the features were passed through a three layer MLP with 512, 512, and seven nodes, with the final features being passed through a hyperbolic tangent nonlinearity.
The model was trained to predict the following seven values: $\Delta x_{\mathbf{c}}$, $\Delta y_{\mathbf{c}}$, $x_{\mathbf{d}}$, $y_{\mathbf{d}}$, $\Delta z_{\mathbf{d}}$, $\cos(\theta)$, and $\text{sign}(\sin(\theta))$, where $\Delta x_{\mathbf{c}}$ is the difference between $x_{\mathbf{c}}$ and the median $x$ value in the extracted point cloud, $\Delta y_{\mathbf{c}}$ is calculated similarly, $\Delta z_{\mathbf{d}}$ is the difference between the height of the bounding box and the height of the human point cloud, and $\theta$ is the heading of the bounding box.
The target values for $\Delta x_{\mathbf{c}}$, $\Delta y_{\mathbf{c}}$, $x_{\mathbf{d}}$, $y_{\mathbf{d}}$, and $\Delta z_{\mathbf{d}}$ were shifted and scaled so that their ranges lied in $[-1, 1]$.
The human-labeled samples were split 97.5/2.5 into training and validation sets, and the model was trained on a single NVIDIA RTX A5500 GPU using the \texttt{AdamW} optimizer \cite{loshchilov2018decoupled} in PyTorch with the default hyperparameters, mean squared error for the loss function, and a batch size of 32.
Every time the validation loss did not improve for five consecutive epochs, we halved the learning rate, and we terminated training if the validation loss did not improve for 10 consecutive epochs.

\section{PointPillars Training Details}
\label{sec:pointpillars}

For all of our PointPillars models, we used the same network hyperparameters described in \citet{lang2019pointpillars} except we used a $100 \times 100$ grid for the \textbf{orchard} task and a $200 \times 200$ grid for the \textbf{urban} task.
For the \textbf{orchard} task, we additionally passed the point coordinates through a NeRF-like positional encoding function \citep{vaswani2017attention, mildenhall2020nerf, tancik2020fourfeat} (the first six of the 10 frequencies used in NeRF) before feeding them into the PointPillars model.
For both the \textbf{orchard} and \textbf{urban} tasks, we applied a point dropping augmentation during training for the \textbf{P2P} models where, for each sample, we sampled a drop probability $\gamma \sim \mathcal{U}(0, 0.2)$ and then dropped points in the object point cloud with probability $\gamma$.
For the \textbf{urban} task, we additionally applied the training augmentations described in \citet{lang2019pointpillars}.
For the \textbf{orchard} task, we applied an additional ``bumping'' training augmentation where, for each sample, we leveled the point cloud using the $\mathbf{R}_{\mathbf{B}}$ and $\mathbf{t}_{\mathbf{B}}$ for the background and then ``unleveled'' the point cloud using the $\mathbf{R}_{\mathbf{B}'}$ and $\mathbf{t}_{\mathbf{B}'}$ for a randomly selected background.
The models were trained on a single NVIDIA RTX A5500 GPU using the \texttt{AdamW} optimizer in PyTorch with the default hyperparameters and a batch size of eight.
Every time the validation loss did not improve for five consecutive epochs, we halved the learning rate, and we terminated training if the validation loss did not improve for 10 consecutive epochs.
For the \textbf{P2P} model, an ``epoch'' was defined as 64,000 training samples for both the \textbf{orchard} and \textbf{urban} tasks.

\begin{figure}[h]
\centering
% Sample 000928.
\includegraphics[scale=0.45]{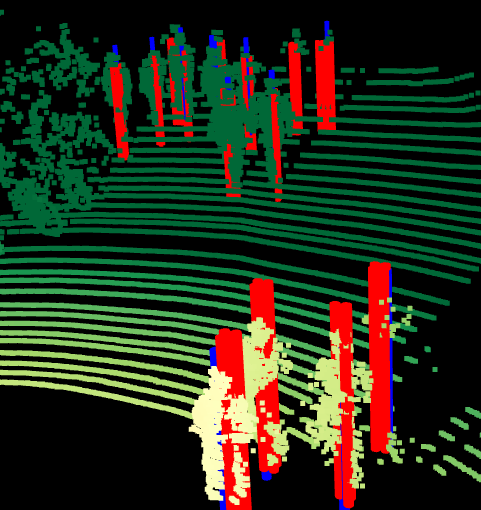}
\caption{A PointPillars model trained on Paved2Paradise data can effectively detect multiple pedestrians in a crowded scene.
The vertical blue bars are the ground truth locations of the bounding box centers for the pedestrians.}
\label{fig:crowd}
\end{figure}

\end{document}